\documentclass[preprint,12pt]{elsarticle}

\usepackage{amssymb}
\usepackage{amsmath}
\usepackage{booktabs}
\usepackage{algorithm}
\usepackage{algpseudocode}
\usepackage{hyperref}
\usepackage[framemethod=tikz]{mdframed}

\begin{document}

\begin{frontmatter}



\title{Early Risk Assessment Model for ICA Timing Strategy in Unstable Angina Patients Using Multi-Modal Machine Learning\tnoteref{tnote1}}%



\author[1,2]{Candi Zheng}\corref{cor1}
\author[3]{Kun Liu}
\author[1]{Yang Wang}
\author[2]{Shiyi Chen}
\author[3]{Hongli Li}
\ead{drhonglili@126.com}
\cortext[cor1]{Corresponding author}
\address[1]{Department of Mathematics,
  Hong Kong University of Science and Technology, Clear Water Bay, Hong Kong SAR, China}
\address[2]{Department of Mechanics and Aerospace Engineering, Southern University of Science and Technology, Xueyuan Rd 1088, Shenzhen, China}
\address[3]{Department of Cardiology, Shanghai General Hospital, Shanghai, China}

\begin{abstract}
\paragraph{Background} Invasive coronary arteriography (ICA) is recognized as the gold standard for diagnosing cardiovascular diseases, including unstable angina (UA). The challenge lies in determining the optimal timing for ICA in UA patients, balancing the need for revascularization in high-risk patients against the potential complications in low-risk ones. Unlike myocardial infarction, UA does not have specific indicators like ST-segment deviation or cardiac enzymes, making risk assessment complex.

\paragraph{Objectives} Our study aims to enhance the early risk assessment for UA patients by utilizing machine learning algorithms. These algorithms can potentially identify patients who would benefit most from ICA by analyzing less specific yet related indicators that are challenging for human physicians to interpret.

\paragraph{Methods} We collected data from 640 UA patients at Shanghai General Hospital, including medical history and electrocardiograms (ECG). Machine learning algorithms were trained using multi-modal demographic characteristics including clinical risk factors, symptoms, biomarker levels, and ECG features extracted by pre-trained neural networks. The goal was to stratify patients based on their revascularization risk. Additionally, we translated our models into applicable and explainable look-up tables through discretization for practical clinical use. 

\paragraph{Results} The study achieved an Area Under the Curve (AUC) of $0.719 \pm 0.065$ in risk stratification, significantly surpassing the widely adopted GRACE score's AUC of $0.579 \pm 0.044$. 

\paragraph{Conclusions} The results suggest that machine learning can provide superior risk stratification for UA patients. This improved stratification could help in balancing the risks, costs, and complications associated with ICA, indicating a potential shift in clinical assessment practices for unstable angina.
\end{abstract}




\end{frontmatter}



\section{Introduction}
Cardiovascular diseases (CVDs) are the leading cause of death globally \cite{Santulli2013EpidemiologyOC}, with ischemic heart disease (IHD) representing the most prevalent and life-threatening type. IHD typically arises from atherosclerotic narrowing within the coronary arteries, which compromises adequate myocardial perfusion. It primarily manifests as either stable ischemic heart disease (chronic coronary syndrome, CCS) or acute coronary syndrome (ACS). Among ACS, the Non-ST-elevation acute coronary syndromes (NSTE-ACS), encompassing conditions such as unstable angina (UA) and Non-ST-elevation myocardial infarction (NSTEMI) \cite{Wiener2013HarrisonsPO}, pose significant diagnostic challenges. This complexity stems from the fact that electrocardiogram (ECG) abnormalities, crucial for ACS diagnosis through the identification of ischemic ST-segment changes, are not always present in NSTE-ACS. The complexity intensifies in the case of UA patients, which spans a wide clinical range from patients with minor arterial narrowing to patients at imminent risk of severe angina or myocardial infarction (MI), necessitating prompt revascularization. Accurately diagnosing UA is essential for determining the optimal timing of such interventions, striving to balance the immediacy of revascularization in high-risk patients against the costs and associated complications.

Invasive procedures, notably coronary arteriography (ICA), are considered the gold standard \cite{10.1093/eurheartj/ehx501.P897} for diagnosing NSTE-ACS by pinpointing atherosclerotic stenosis within the coronary arteries. However, ICA is expensive and is accompanied by inherent risks of complications ranging from short-term sequelae to life-threatening damages \cite{Tavakol2012RisksAC}. As a result, though the potential diagnostic benefit of ICA is irrefutable, the timing of performing ICA is controversial. After the initial medical treatment of UA, two main strategies are typically weighed: the routine invasive strategy and the selective invasive strategy \cite{Collet20202020EGstrategy}. The former recommends ICA for all patients, while the latter suggests reserving ICA for cases with evidence of recurrent ischemia. Studies indicate that under the routine strategy, approximately two-thirds of patients undergo unnecessary ICA when no significant stenosis is present \cite{Patel2010LowDY}, prompting the selective strategy to minimize both the financial burden and the risk of complications by avoiding needless procedures. However, it's noteworthy that the selective invasive strategy does not equal the routine strategy in reducing high-risk events such as myocardial infarction (MI), severe angina, and rehospitalization \cite{Mehta2005RoutineVS}. To effectively balance the risks, costs, and complications associated with ICA, a robust early risk assessment is crucial to identify patients who would most benefit from this intervention.

Early risk assessments for ICA require risk assessment models based on timely accessible and available data, such as demographic characteristics, clinical risk factors, symptoms, and biomarkers levels. Several models have been developed to estimate mortality or myocardial infarction risk in this context \cite{Goldman1996PredictionOT, Antman2000TheTR, Sanchis2005NewRS, Fox2006PredictionOR}. Notably, the Grace risk score is distinguished by its superior discriminative performance \cite{Fox2006PredictionOR, Collet20202020EGgrace}. However, these models predominantly focus on mortality and are tailored for a broad spectrum of acute coronary syndrome (ACS) patients, encompassing both angina and myocardial infarction. A significant limitation is their reliance on ST-segment deviations in ECG readings as indicators of ischemia, rendering them less effective for UA patients who typically present lower mortality risk and lack characteristic ST-segment changes. Moreover, machine learning models designed for predicting cardiovascular disease/stroke risks or major adverse cardiovascular events \cite{Jamthikar2020MulticlassML, Jamthikar2020CardiovascularstrokeRP, Kakadiaris2018MachineLO, Stewart2021ApplicationsOM} are oriented towards minimizing adverse events or death. Although there are machine learning models aimed at assisting diagnosis \cite{Qin2022MachineLA, Stewart2021ApplicationsOM}, they offer limited guidance for ICA decision-making. Consequently, there remains a gap in the development of specific risk stratification models for ICA timing of UA patients that effectively balance the risks, costs, and complications.

In this study, we developed multi-modal machine learning-based models to assess the optimal timing for invasive coronary arteriography (ICA) in unstable angina (UA) patients. The model is trained on multi-modal data consists of regular text-based medical history and multi-channel electrocardiograms (ECG), collected from 640 UA patients in Shanghai General Hospital. These data were processed into features encompassing demographic characteristics, clinical risk factors, symptoms, biomarker levels, and ECG characteristics, the latter extracted using pre-trained neural networks. Patients were categorized into low-risk and high-risk groups based on whether they underwent revascularization. We employed linear logistic regression (LR) and non-linear gradient boosted tree (GBDT) models to stratify patients by their risk of requiring revascularization. Both models outperformed the widely used GRACE score. However, the enhancement in performance offered by the non-linear GBDT model and multi-modal ECG information was not statistically significant. These results suggest that enhanced risk stratification can be achieved by more effectively utilizing conventional medical history data. This advancement has the potential to improve the decision-making process regarding the necessity, timing, and risks associated with ICA in UA patients, thereby optimizing patient outcomes while considering healthcare resource allocation.

\section{Material and Methods}

\subsection{Participants}
The clinical data for this study were sourced from the Department of Cardiology at Shanghai General Hospital. We recorded data from 640 patients diagnosed with unstable angina between September 2017 and August 2021. The inclusion criteria for the study were as follows: (1) Diagnosis of unstable angina; (2) A history of undergoing coronary arteriography. Conversely, patients were excluded based on any of the following criteria: (1) History of percutaneous coronary intervention (PCI) and pacemaker implantation; (2) Severe electrocardiogram (ECG) abnormalities such as bundle branch block or ST-T changes not attributable to coronary artery diseases; (3) A history of or current diagnosis of myocardial infarction; (4) Chest pain resulting from non-coronary artery diseases; (5) Diagnosis of stable angina; (6) Refusal of PCI treatment; (7) Clinical indications of heart failure.

From the collected data, 640 patients with unstable angina who underwent invasive coronary arteriography (ICA) were analyzed. Of these, 284 patients received PCI revascularization, while the remainder were managed with conservative medical treatment. All patients were classified under Killip class I, indicating no clinical evidence of heart failure. Among them, 150 patients were noted to have suspicious ST-T segment deviations—elevation, depression, or other changes—as mentioned in their diagnostic reports. Additionally, their GRACE scores were computed in accordance with the guidelines \cite{Collet20202020EGgrace}.

\subsection{Processing of Raw Clinical Data}
The original dataset comprised patients' medical histories, laboratory test results, and electrocardiograms (ECGs) stored as JPG files. The medical histories and laboratory test data were digitized using optical character recognition (OCR) technology based on convolutional neural networks. This process was followed by manual proofreading and error correction to ensure accuracy. Clinical features were then automatically extracted from the digitized medical history texts via keyword matching. We addressed errors arising from transcription or OCR inaccuracies through a combination of manual checks and automatic outlier detection methods. The K-nearest neighbors (KNN) algorithm \cite{Angiulli2002FastOD} was utilized to identify transcription errors, such as missing decimals or digits. Histogram analysis was conducted to manually identify systematic extraction errors, indicated by irregular spikes in the data.

All identified errors and outliers were subsequently rectified through manual intervention. Missing values were manually verified and filled if overlooked during automatic extraction; otherwise, they were assigned a placeholder value of $-1$. The ECG data were digitized by extracting signal pixels from the JPG files. This was accomplished using a signal search agent combined with the Monte-Carlo tree search algorithm. We processed ECG signal segments of 2.5 seconds at a sampling rate of 57 Hz. Additionally, we adjusted the amplitude scale according to the resampling rate. This ensured that signals extracted from ECGs of varying resolutions were standardized to a consistent voltage scale.

\subsection{Clinical Features and Targets}
The primary objective of our machine learning models is the early risk assessment for invasive coronary arteriography (ICA) in patients with unstable angina (UA). Specifically, we focus on modeling the rate of revascularization, which represents the proportion of patients undergoing ICA that subsequently receive coronary revascularization. Coronary revascularization procedures, such as percutaneous coronary intervention (PCI), are critical for restoring blood flow in occluded coronary arteries, indicating severe arterial stenosis requiring immediate intervention. Thus, the revascularization rate serves as a reliable indicator of a patient’s risk level and the necessity of ICA. In practice, patients were classified into low-risk and high-risk groups based on their revascularization status. Our model aims to stratify patients into these risk categories by predicting their likelihood of requiring revascularization.

\begin{table}[h]
\scalebox{0.8}{
\begin{tabular}{@{}llllll@{}}
Clinical Features & Feature Type    & Feature Unit      & Mean Value & SD$^h$ & $\%$ Missing \\ \midrule
Age       & Numerical & Years Old & 65.4      & 9.3 & 0.0\\
BMI      & Numerical & $\mbox{Kg}/\mbox{m}^2$ & 24.6      & 3.5 & 7.8 \\
Gender    & Nominal & Male; Female &   -     &  \\
Heart rate& Numerical &  beats/min &  81.2  &12.6 & 3.6\\
Systolic blood pressure& Numerical &  mmHg &  137.2  &20.3 & 0.0\\
Diastolic blood pressure& Numerical &  mmHg &  79.3  & 11.7 & 0.0\\
Smoking &  Nominal &  Positive/Negative  & -\\
Diabetes&  Nominal &  Positive/Negative  & -\\
Hypertension&  Nominal &  Positive/Negative  & -\\
ST deviation&  Nominal &  Positive/Negative  & -\\
Glycerin tricaproate &  Numerical &  mmol/L  & 1.6 & 1.1 & 8.0\\
LDL$^a$ &  Numerical &  mmol/L  & 2.9 & 1.6 & 6.9\\
HDL$^b$ &  Numerical &  mmol/L  & 1.1 &0.3 & 11.4\\
CKMB$^c$ &  Numerical &  ng/ml  & 2.8 & 9.8 & 2.6\\
BNP$^d$ &  Numerical &  pg/ml  & 91.5 & 268.5 & 7.5\\
D-Dimer &  Numerical &  mg/L  & 0.47 & 0.72 & 18.8\\
Myoglobin&  Numerical &  ng/ml  & 39.9 & 172.1 & 1.9\\
CRP$^e$&  Numerical &  mg/L  & 3.3 & 8.8 & 20.1 \\
TnI$^f$&  Numerical &  ng/ml  & 0.1 &1.3&5.5\\
SCR$^g$&  Numerical &  umol/L  & 77.0 & 81.4 & 7.5\\
\end{tabular}}
\caption{The clinical features used in our early risk assessment model}
\footnotesize{$^a$ Low-density lipoprotein Cholesterol $^b$ High-density lipoprotein Cholesterol $^c$ Creatine Kinase-MB $^d$ B-Type natriuretic peptide $^e$ C-reactive protein $^f$ Troponin I $^g$ Serum Creatinine $^h$Standard Deviation }\\
\label{tab:1}
\end{table}

The clinical features utilized in our model, as detailed in Table \ref{tab:1}, encompass demographic details, clinical risk factors, symptoms, and biomarker levels. These features were processed prior to inputting into the machine learning algorithms. Categorical features were transformed into binary integers via one-hot encoding. Non-negative numerical features, particularly those with wide-ranging values (such as BNP, CKMB, Creatinine, D-Dimer, Glycerin tricaproate, LDL, HDL, Myoglobin, CRP, and Tn-I), were log-transformed to normalize their ranges. For logistic regression, missing values were imputed with the median of the respective feature, while for gradient boosted trees, missing values were left as is since these models can inherently handle them.

In addition to the clinical features, original 12-channel ECG signals were incorporated into our model. ECG signals were processed into 10-second segments, each with 1000 sample points per channel. Given the high correlation between these sample points, direct use as inputs for machine learning algorithms was not feasible. Instead, we utilized neural networks pre-trained on the PTB-XL dataset \cite{Wagner:2020PTBXL, Wagner2020:ptbxlphysionet, Goldberger2020:physionet} for signal processing. The ECG signals, resized to (12,1000), were input into the Inception-1D neural network \cite{Strodthoff2021DeepLF}, with the final layer removed. This process reduced the dimensionality of ECG signals to 128, making them suitable as inputs for subsequent machine learning algorithms.

\subsection{Machine Learning Models}

In this study, we concentrate on two distinct machine learning algorithms: the non-linear Gradient-Boosted Decision Tree (GBDT) and the linear Logistic Regression (LR). The GBDT, a sophisticated non-linear model, operates by integrating multiple decision trees within a gradient boosting framework. This method significantly enhances performance over single decision trees. XGBoost stands out as an renowned variant of GBDT for its accelerated training speed and superior generalization capabilities in comparison to the classic GBDT. We employ the XGBoost package as our primary tool for modeling revascularization risk.

On the other hand, Logistic Regression represents the archetype of linear models in medical data analysis. Its widespread use is attributed to its simplicity and interpretability, yielding models that facilitate detailed statistical analysis and inference. Therefore, logistic regression was chosen as our secondary algorithm for risk assessment.

In summary, our approach utilizes both the non-linear GBDT and the linear Logistic Regression model. This combination of methodologies allows compare non-linear and linear modeling techniques in medical data analysis, yielding a comprehensive and robust assessment of revascularization risk.

\subsection{Feature selection}
In our analysis, we initially identified 148 features, comprising 20 clinical features and 128 ECG-derived features. However, it is crucial to acknowledge that not all of these features are equally informative for risk assessment. Features that do not correlate significantly with the revascularization rate can introduce unnecessary noise and potentially degrade the performance of our risk assessment model. Consequently, a meticulous feature selection process is essential to filter out those features that are most relevant and correlated with the revascularization rate.

Our feature selection strategy encompasses three distinct approaches. The first is Forward Sequential Feature Selection \cite{Ferria2007ComparativeSO}, a method that incrementally selects features based on their ability to enhance the cross-validation Area Under the Curve (AUC) results. This process is guided by a minimum improvement threshold of 0.03. 

The second method we employed is univariate Analysis of Variance (ANOVA) based on the F-test. The univariate F-test compares the variance explained with those unexplained for each feature. A feature that explains more variance will contribute more to the risk assessment and have a higher F-value. Features with significantly F-values larger than a threshold will be selected, with the threshold determined according to the Benjamini-Hochberg Procedure at a false discovery rate of 0.05 \cite{Efron2016ComputerAS}. Prior to this selection, missing values are imputed using the mean of the respective features. 

The third approach is multivariate ANOVA, which also relies on mean imputation for missing values and assesses feature relevance through the T-test. Instead of comparing each feature, the multivariate T-test fits a logistic regression model on all features and detects features with non-zero coefficients. Features with more significant coefficients contribute more to the logistic regression model and hence have larger T-values and lower p-values. We hence select features with large T-values whose p-value is less than 0.05. The Benjamini-Hochberg procedure is not adopted here because these T-tests are not independent. 

By implementing these three feature selection methods, we aim to refine our dataset, ensuring that our risk assessment model is both efficient and effective, utilizing only the most informative features in predicting the revascularization rate.

\subsection{Hyper-parameters of Machine Learning Models}
We determine the hyper-parameters of the machine learning model using the grid search method. The grid search method determines the optimal hyperparameter over a parameter grid spaned by all possible combinations of provided parameters. The optimal hyperparameter is the combination yielding the highest area-under-curve (AUC) score computed via cross-validation. For the gradient boosted tree, we provide the grid search method with three parameters to optimize, namely the learning rate (0.02,0.11,0.2), subsample (0.5,0.75,1.0), and the number of estimators (6,12,24). As for other parameters, we use the default parameters of the XGBoost package except for maximum tree depth 2, alpha 2.75, and gamma 2.75. For the linear regression, we use L1 regularization with three choices of the parameter C (0.1, 0.5, 1). We emphasize that the grid search is integrated as a part of the machine learning model, with the cross-validation performed on the training data. Therefore there is no data leakage from the test set which causes optimally biased performance.

\subsection{Validation of the Model Performance}

We evaluate the performance of our machine learning model using the receiver operator characteristic (ROC) curve and its area under the curve (AUC) \cite{Bradley1997TheUO}. The ROC curve plots the model's true-positive rate against the false-positive rate at various threshold values. It demonstrates the model's ability at various threshold values to distinguish patients requiring revascularization from those who do not. The AUC summarizes the ROC curve into a single value. It also measures the probability of the patient requiring revascularization receiving a greater score than those who do not. The ROC curve and AUC are considered statistically more consistent and more discriminating than classification accuracy \cite{Ling2003AUCAS}. Moreover, it is well-defined for unbalanced data sets. Therefore we compare the ROC curve and AUC among our machine learning models and the traditional GRACE score to demonstrate their risk assessment power.

We estimate the generalization ability of our model's performance on previously unseen data. In practice, we split the data set into training and test parts. The machine learning models are trained on training data. In contrast, the ROC curve and AUC score were computed on the test data. Moreover, in order to obtain confidence intervals, we use the five-fold stratified cross-validation, which tests the model performance on five different train-test splitting configurations. The merit of cross-validation is that it yields several AUC score estimations, allowing us to compute its confidence interval. However, the cross-validation result could be biased by data leakage if we tune the model's parameters w.r.t the cross-validation result, which contains information from both training and test data. 

To avoid such data leakage, inside each cross-validation fold, we perform another cross-validation process for parameter tunning that solely uses the training part of each train-test splitting configuration. This helps us to avoid tuning parameters on test data hence avoid possible data leakage when computing the ROC curve and AUC score. In summary, we use two nested cross-validation processes, one for hyper-parameter tunning and one for performance evaluation to compute the ROC curve and AUC score.

\subsection{Determining Thresholds and Probability Calibration}

In our study, machine learning algorithms assign a risk score ranging from 0 to 1 for each patient. To make informed decisions regarding invasive coronary arteriography (ICA), it's necessary to convert these risk scores into decision using various thresholds corresponding to different true-positive rates. However, directly employing thresholds derived from previously cross-validated Receiver Operating Characteristic (ROC) curves is not feasible, as this could lead to data leakage due to the inclusion of information from the test sets. To circumvent this issue, we establish thresholds using cross-validation solely within the training set. We conduct cross-validation of our model on this training set to generate an ROC curve and identify the associated thresholds. These thresholds are then assessed on the test set to determine their respective true and false-positive rates. Notably, this cross-validation process is conducted in parallel with hyper-parameter tuning.

Once the true-positive and false-positive rates at certain threshold $r$ are determined, we can compute a certain patient's revascularization probability $P(\mbox{Rev}|\mbox{risk score} \ge r)$ given his/her risk score $r$ predicted by a machine learning algorithm as

\begin{equation}
    P(\mbox{Rev}|\mbox{risk score} \ge r) = \frac{\mbox{TPR} \times P(Rev)}{\mbox{TPR} \times P(Rev)+\mbox{FPR} \times P(Not Rev)},
\end{equation}
in which $\mbox{TPR}$ and $\mbox{FPR}$ represents the true/false positive rate, $P(Rev)$ and $P(Not Rev)$ are the probability of patients in our dataset requiring/not requiring revascularization.

\subsection{Explainable Machine Learning: Risk Stratification Look-Up Table}

Explainable machine learning models, especially when presented as risk stratification look-up tables, offer significant advantages for physicians over complex computer-based algorithms. These models enhance transparency, facilitate understanding of decision-making processes, and do not require extensive training or computational resources. To harness these benefits, we have transformed our logistic regression and GBDT models into interpretable look-up tables through discretization, while excluding computationally intensive pre-trained ECG features reliant on neural networks.

The transformation of the GBDT model into an interpretable format is straightforward. The risk score in the GBDT model is derived from the sum of scores across various leaf nodes of multiple decision trees. These trees pose questions like "Is the gender male?" or "Is the CKMB level below a certain threshold?", with each answer leading to different leaf node scores. For practical use, we have transcribed these decision-making questions and their corresponding leaf node scores into a user-friendly format. This approach allows healthcare professionals to easily calculate risk scores by adding up the scores from the relevant leaf nodes. Examples of these look-up tables are provided in the appendix, as shown in Fig\ref{fig:1t} and Fig\ref{fig:2t}.

Simplifying the logistic regression model into a look-up table involves discretizing continuous clinical features. The original model calculates risk scores by summing the weighted values of each feature. While discrete features like gender are straightforward to represent, continuous biomarker values require quantile-based discretization. We have evenly spaced these values into quantiles, assigning corresponding weighted values in the look-up table. This method approximates continuous values to their nearest quantiles, enabling physicians to easily calculate the final risk score by summing weighted values from the table. The details of this discretization and the resultant look-up table are also included in the appendix, as depicted in Fig\ref{fig:3t} and subsequent figures. These look-up tables make our machine learning models not only more accessible to physicians but also more transparent and understandable, enhancing their applicability in clinical practice.

\section{Results}

\begin{figure}[h!] 
\begin{center}
\includegraphics[width=14cm]{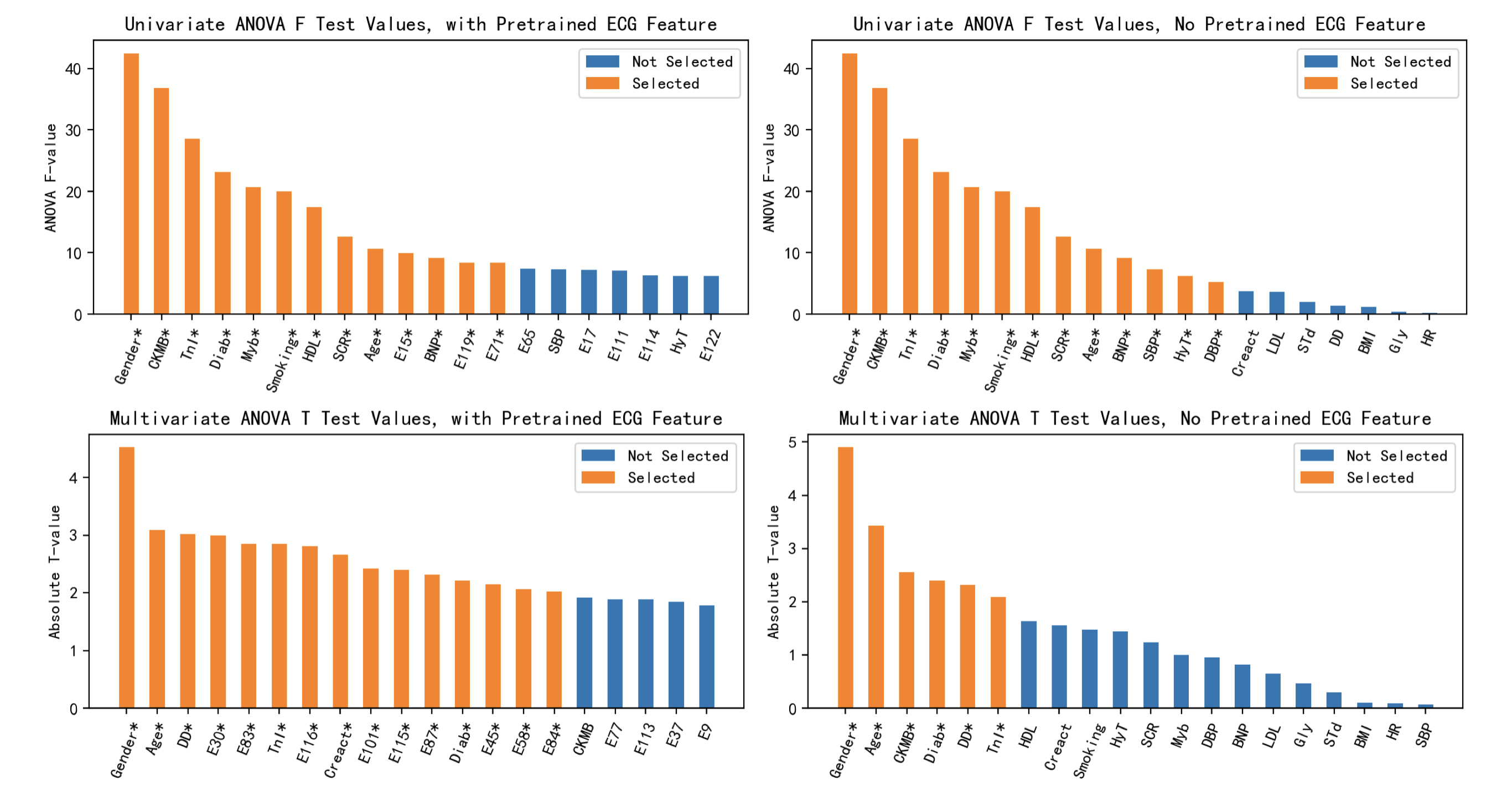}
\end{center}
\caption{ The ANOVA analysis of clinical and pre-trained ECG features. We use both univariate F-test and multivariate T-test to demonstrate the contribution of each feature. For the univariate F-test, we select features according to the Benjamini-Hochberg Procedure with the false discovery rate controlled at 0.05. For the multivariate T-test, we select features with p-values under 0.05. LDL: Low-density lipoprotein Cholesterol; HDL: High-density lipoprotein Cholesterol; CKMB: Creatine Kinase-MB; BNP: B-Type natriuretic peptide, CRP C-reactive protein; TnI: Troponin I; SCR: Serum Creatinine; Diab: Diabetes; HyT: Hyper-tension; STd: ST-segment deviation; Gly: Glycerin tricaproate; HR: Heart rate; SBP: Systolic blood pressure; DBP: Diastolic blood pressure; BMI: Body mass index; E: Pretrained ECG feature labeled from 0 to 127; }\label{fig:1}
\end{figure}

We demonstrate the contribution of each feature using ANOVA analysis based on univariate F-test and multivariate T-test. In the univariate F-test analysis, features were selected based on their F-values and the Benjamini-Hochberg procedure with a controlled false discovery rate of 0.05. The selected features demonstrate a significantly higher F-value, suggesting a greater variance explained by these features in the model. Features like gender, chest pain type (CPK), diabetes (Diab), and myocardial infarction biomarker troponin I (TnI) show high F-values, indicating their strong association with the outcome variable, which in this context is the revascularization rate.

the ANOVA without pretrained ECG features presented in Figure \ref{fig:1} underscores the significance of various clinical and biochemical parameters in stratifying the risk of revascularization. Notably, traditional risk factors such as the presence of diabetes and a history of smoking show high F-values hence have strong associations with an increased likelihood of revascularization. Cardiac biomarkers, specifically Creatine Kinase-MB (CKMB), Myoglobin, D-Dimer, and Troponin I (TnI), are also identified as significant contributors to the risk assessment model, aligning with their clinical relevance in indicating myocardial damage.

Furthermore, the ANOVA with pretrained ECG features elucidates the role of multi-modal ECG information. Notably, several pre-trained ECG features (E0 to E127) have been selected due to their significant T-values, pointing towards their predictive power in the model. Interpreting their statistical significance is a great challenge due to the "black-box" nature of neural networks. Howver, it is evident from the figure that the inclusion of pre-trained ECG features alters the landscape of selected clinical features. For example, with ECG features excluded, biomarkers such as CKMB are selected by multivariate ANOVA, which were not selected when ECG features were included. Such changes are potential hints of the correlation between ECG information and biochemical parameters.

Additionally, the analysis indicates a reduced emphasis on heart rate and ST-segment changes, which may reflect the tailored approach of our model to the UA population, wherein such indicators may not be as pronounced as in patients with more acute ischemic presentations. Overall, the insights from Figure \ref{fig:1} suggest a nuanced approach to risk stratification that is contextually informed by the pathophysiology of the UA patient group.

\begin{table}[h]
\scalebox{0.9}{
\begin{tabular}{@{}llllll@{}}
Model/Feature & AUC Without Pretrained ECG   & AUC With Pretrained ECG \\ \midrule
Logistic Regression       & $0.719 \pm 0.065^{*}$ & $0.712 \pm 0.078$\\
Gradient Boosted Tree       & $0.710 \pm 0.081$ & $0.689 \pm 0.041^{**}$\\
GRACE       & $0.579 \pm 0.044$ & -    \\
\end{tabular}}

\footnotesize{ The uncertainty is +- 1.98$\times$standard deviation with 95\% confidence. $^{*}$ Greater than the GRACE score, $p<0.05$, one-side t-test with unequal variances. $^{**}$ Greater than the GRACE score, $p<0.01$, one-side t-test with unequal variances }\\
\caption{The AUC score of the machine learning models is compared with the GRACE score. The confidence interval of logistic regression and GBDT are evaluated via standard deviation of cross-validation results. The confidence interval of the GRACE score is evaluated via bootstrap. Machine learning algorithms have better AUC than the GRACE score. However, there are no significant performance differences between machine learning algorithms.}
\label{tab:1}
\end{table}

\begin{figure}[h!] 
\begin{center}
\includegraphics[width=14cm]{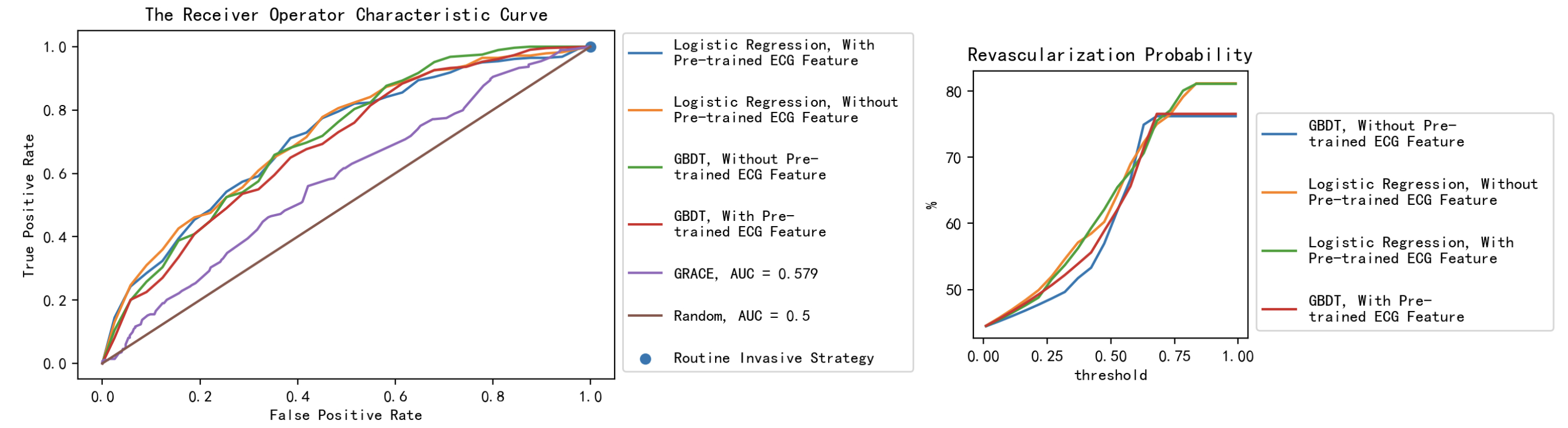}
\end{center}
\caption{The ROC curve of machine learning models and the corresponding revascularization probability for each classification threshold. Machine learning models evaluate patients' risk as scores in the range (0,1). Patients with a score higher than a specific threshold value are considered high-risk (Requiring revascularization). The ROC curve plots the model's true-positive rate against the false-positive rate at various threshold values. Both true-positive and false-negative rates are computed via nested cross-validation to prevent data leakage from the test set. The revascularization probability states the probability of revascularization given a patient with a risk score larger than the threshold. It demonstrates the discrimination power of machine learning models. Note that all machine learning models have a lower false-positive rate than the GRACE score given a true-positive rate. This means our model could reduce the false-positive rate hence avoiding unnecessary ICA.}\label{fig:2}
\end{figure}

The Receiver Operating Characteristic (ROC) curve, depicted in Figure \ref{fig:2}, illustrates the performance of our machine learning model in stratifying the risk of revascularization for unstable angina (UA) patients. Under a routine invasive strategy, all UA patients are recommended to undergo invasive coronary arteriography (ICA), which corresponds to a true-positive rate (TPR) of 1. This strategy ensures that all high-risk patients are identified for ICA but results in a false-positive rate (FPR) of 1, signifying that many low-risk patients also receive unnecessary ICA.

Our machine learning approach proposes a selective invasive strategy based on assessed risk scores, which aims to reduce these unnecessary procedures. The ROC curve demonstrates that by adjusting the TPR threshold to 0.9, we can substantially decrease the FPR to 0.7, which is $22\%$ better than routine invasive strategy and $12\%$ better than GRACE score. This adjustment indicates that fewer low-risk patients are subjected to ICA, thus avoiding unwarranted interventions.

Consequently, the application of our risk assessment model presents a significant opportunity to mitigate healthcare costs and minimize patient exposure to potential complications associated with unnecessary ICA. The ability to maintain a reasonable TPR while significantly reducing the FPR underscores the efficacy of our machine learning model in refining the decision-making process for ICA in the context of UA patient care.

Table \ref{tab:1} details the area under the receiver operator characteristic curve (AUC) for our evaluated models. Each model, including both logistic regression and gradient-boosted decision tree (GBDT) variants, demonstrates superior AUC performance when compared to the GRACE score. Notably, the logistic regression model excluding pre-trained ECG features and the GBDT model incorporating these features display statistically significant improvements. This enhancement also suggests that the non-linear GBDT model derives greater benefit from the inclusion of multi-modal ECG features than the linear logistic regression model, potentially due to the GBDT model's enhanced resilience to the collinearity often present in ECG data.

However, the limited size of our dataset precludes a definitive analysis of the individual impact of ECG features. Consequently, more comprehensive research with a larger dataset is required to precisely evaluate the contribution of pre-trained ECG features to risk assessment models and confirm their utility in clinical settings.

\section{Discussion}
Our risk assessment model has the potential to curtail both the financial burden and the risk of complications by minimizing the incidence of unnecessary invasive coronary arteriography (ICA). The logistic regression and gradient-boosted decision tree (GBDT) models we developed both report AUC values around 0.7, outperforming the GRACE score, which has an AUC of 0.579. According to the ROC curve analysis, a routine invasive strategy, which recommends ICA for all patients, inherently yields true-positive and false-positive rates of 1. However, by adopting a selective invasive strategy that utilizes our model's risk scores, we can decrease the false-positive rate significantly from 1 to approximately 0.7 by adjusting the true-positive rate from 1 to 0.9. This equates to a $22\%$ improvement over the routine strategy and a $12\%$ improvement over the GRACE score in averting unnecessary ICA procedures. In essence, our machine learning-based risk assessment demonstrates an enhanced ability to finely tune the balance between the risk and the benefits of ICA, leading to a more efficient use of healthcare resources and better patient care outcomes.

Our study underscores the critical role of certain risk factors and cardiac enzymes, namely diabetes, , gender, Creatine Kinase-MB (CKMB), and Troponin I (Tn-I), etc, in the invasive coronary arteriography (ICA) risk assessment for patients with unstable angina (UA). The detailed weights for these risk factors are demonstrated in look-up tables in Fig\ref{fig:1t}, Fig\ref{fig:2t}, and Fig\ref{fig:3t}. Conversely, several classical indicators such as heart rate and ST-segment deviation, typically significant in the context of ischemic events, were found to be less relevant. This may be attributable to the transient nature of ischemia in non-ST elevation acute coronary syndromes (NSTE-ACS), particularly in UA patients, which renders these ECG indicators less reliable for diagnosis. Consequently, the GRACE score, which heavily weighs heart rate and ST-segment deviation, appears less effective for this patient group. In contrast, by prioritizing the aforementioned risk factors and cardiac enzymes, our machine learning models have achieved AUC scores that significantly surpass those of the GRACE score, indicating a more robust and relevant tool for risk stratification in this context.

ECG features beyond ST-T segments may offer valuable insights for risk stratification in unstable angina (UA) patients. We utilized inception-1D neural networks to extract pre-trained ECG features from the extensive PTB-XL dataset, which includes records from 21,837 patients. These pre-trained features have demonstrated their effectiveness in predicting various diagnostic categories and patient demographics, as evidenced in previous studies \cite{Strodthoff2021DeepLF}. Our analysis also reveals that these ECG features, within both univariate and multivariate feature selection frameworks, exhibit significant predictive capabilities that surpass those of ST-segment deviation, as depicted in Figure \ref{fig:1}. Nevertheless, the size of our dataset limits our capacity to fully discern their individual effects on model performance when contrasted with models developed exclusively on risk factors and cardiac enzymes. To definitively ascertain the potential contribution of ECG features to ICA risk assessment, further research with a larger dataset, capable of providing greater statistical resolution, is necessary.

The logistic regression algorithm demonstrates satisfactory performance in risk stratification for invasive coronary arteriography (ICA), offering a robust linear approach. While the non-linear gradient-boosted decision tree (GBDT) algorithm is generally favored for its ability to capture complex, non-linear relationships, in our study, it did not produce an AUC score that was significantly superior to that of the logistic regression model. This finding aligns with prior research indicating that non-linear machine learning algorithms do not always yield performance improvements in the medical domain \cite{Christodoulou2019ASR}. Notably, we observed that incorporating pre-trained ECG features into the GBDT model resulted in a reduced variance and a significantly improved performance compared to the GRACE score. This suggests a synergistic effect between the GBDT algorithm and pre-trained ECG features, hinting at the possibility that this combination could enhance model performance. Consequently, the potential for performance gains using a mix of non-linear GBDT models and pre-trained ECG features remains an open avenue for future exploration.

A notable limitation of our risk assessment model lies in defining its inclusion criteria. Our study specifically targets patients with unstable angina (UA), deliberately excluding those with other forms of acute coronary syndrome (ACS), particularly those displaying prominent ischemic indicators such as elevated heart rates, ST-T segment changes, or cardiac enzymes associated with myocardial infarction. Consequently, our model is more finely tuned to discern the revascularization risk in UA patients, rather than identifying more severe ischemic symptoms found in other ACS categories. This focus necessitates that physicians perform a thorough and thoughtful initial screening for ischemic indicators before applying our model for risk assessment. This screening process, however, is inherently subjective and may vary among clinicians. Therefore, further research is needed to develop a more objective approach to standardize this preliminary filtration process, enhancing the model's clinical utility and consistency across different practitioners.

In summary, our study introduces a machine learning-based selective invasive strategy for determining the timing of invasive coronary arteriography (ICA) in unstable angina (UA) patients. Given that UA patients have a lower mortality risk than those with myocardial infarction, a selective approach is vital for balancing the risks, costs, and potential complications of intervention. Traditional tools like the GRACE score, focused on higher-risk ACS patients, lack specificity for UA. Our research fills this gap with tailored machine learning models for UA patients, yielding significantly higher AUC scores than the GRACE score and thereby offering more precise and reliable risk stratification for this group.

\section*{Acknowledgments}

\appendix

\begin{figure}[h!]
\begin{center}
\includegraphics[width=15cm]{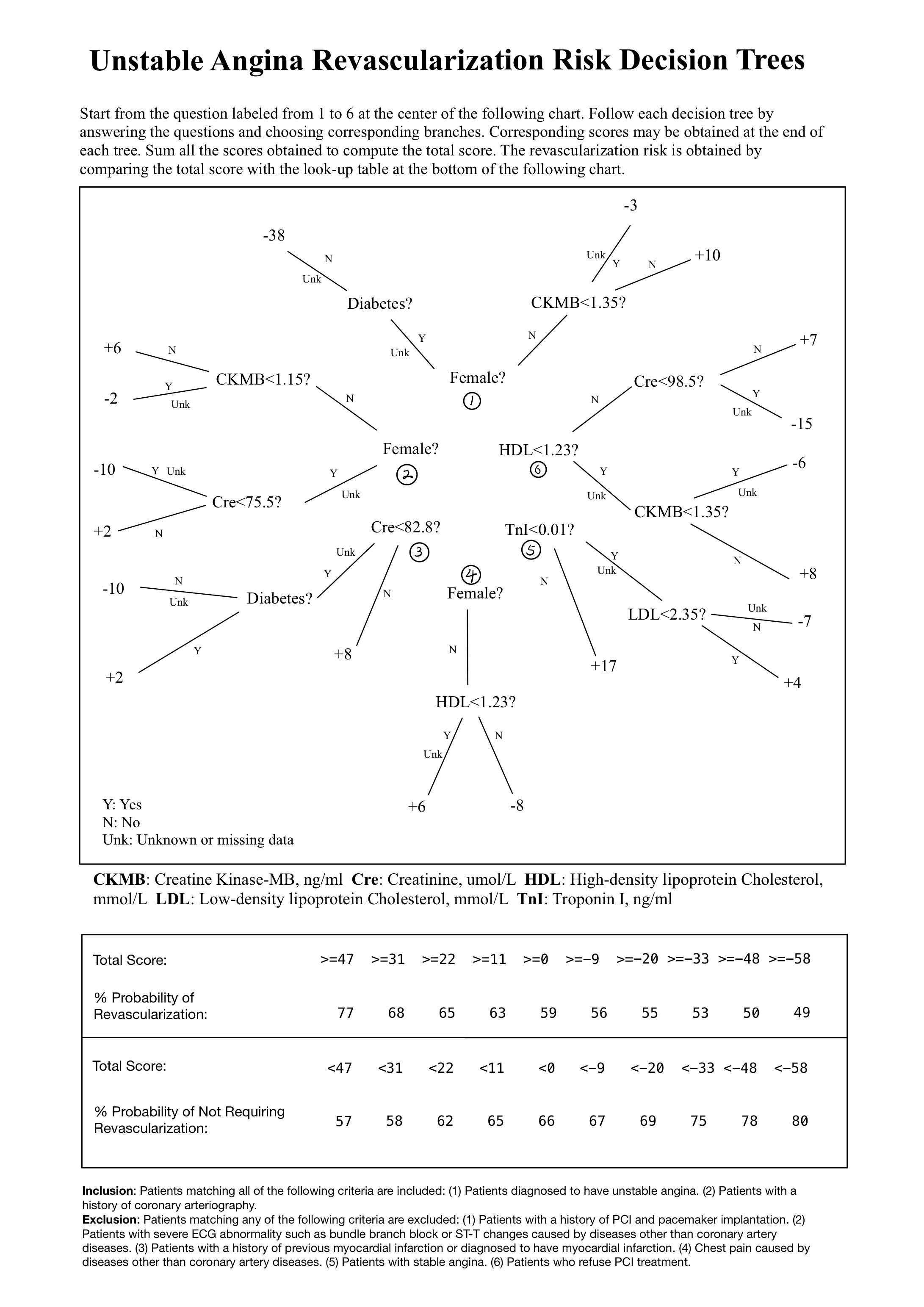}
\end{center}
\caption{Risk scores look-up table Derived from the GBDT model}\label{fig:1t}
\end{figure}
\begin{figure}[h!]
\begin{center}
\includegraphics[width=15cm]{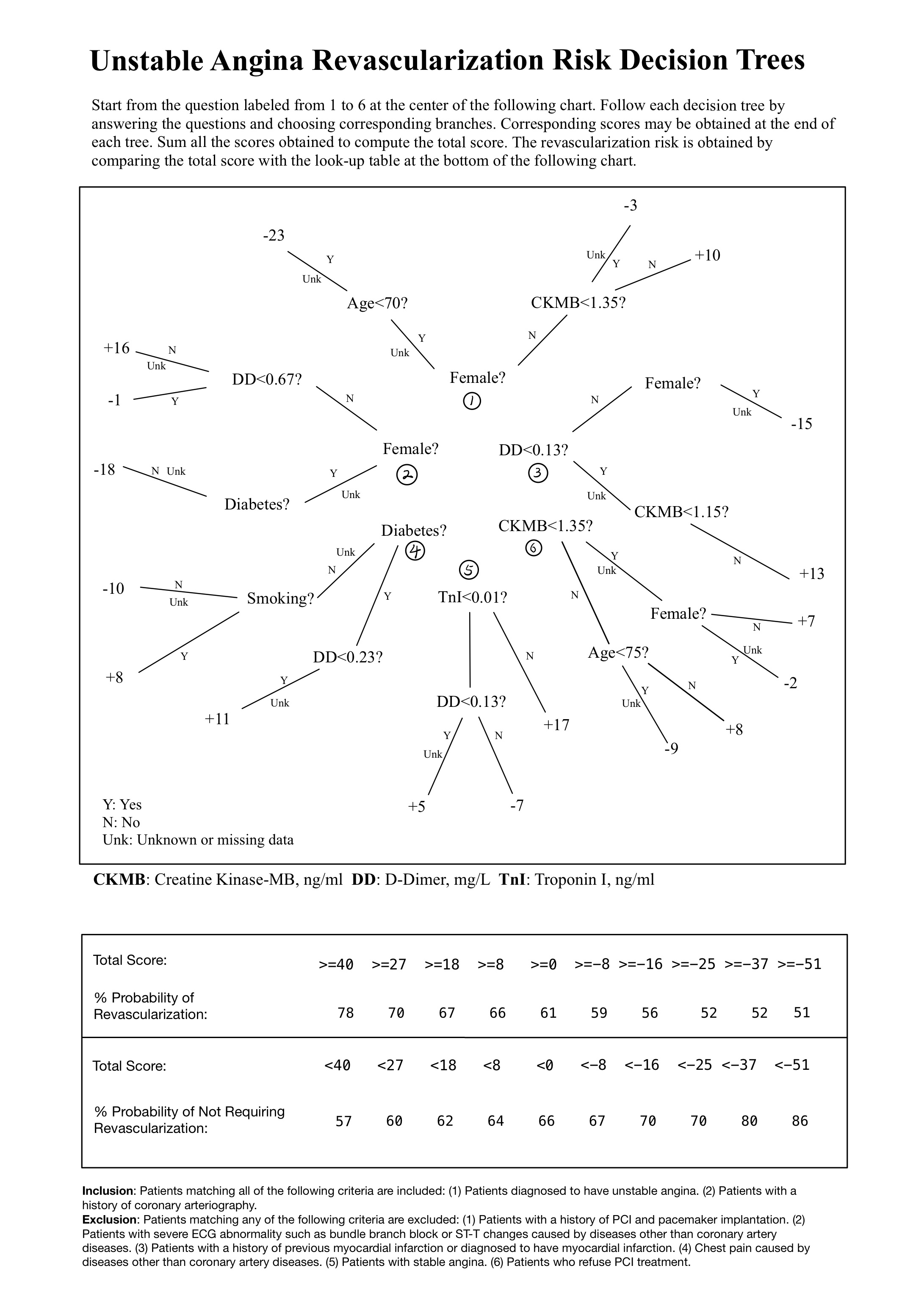}
\end{center}
\caption{Another Risk scores look-up table Derived from the GBDT model}\label{fig:2t}
\end{figure}

\begin{figure}[h!]
\begin{center}
\includegraphics[width=15cm]{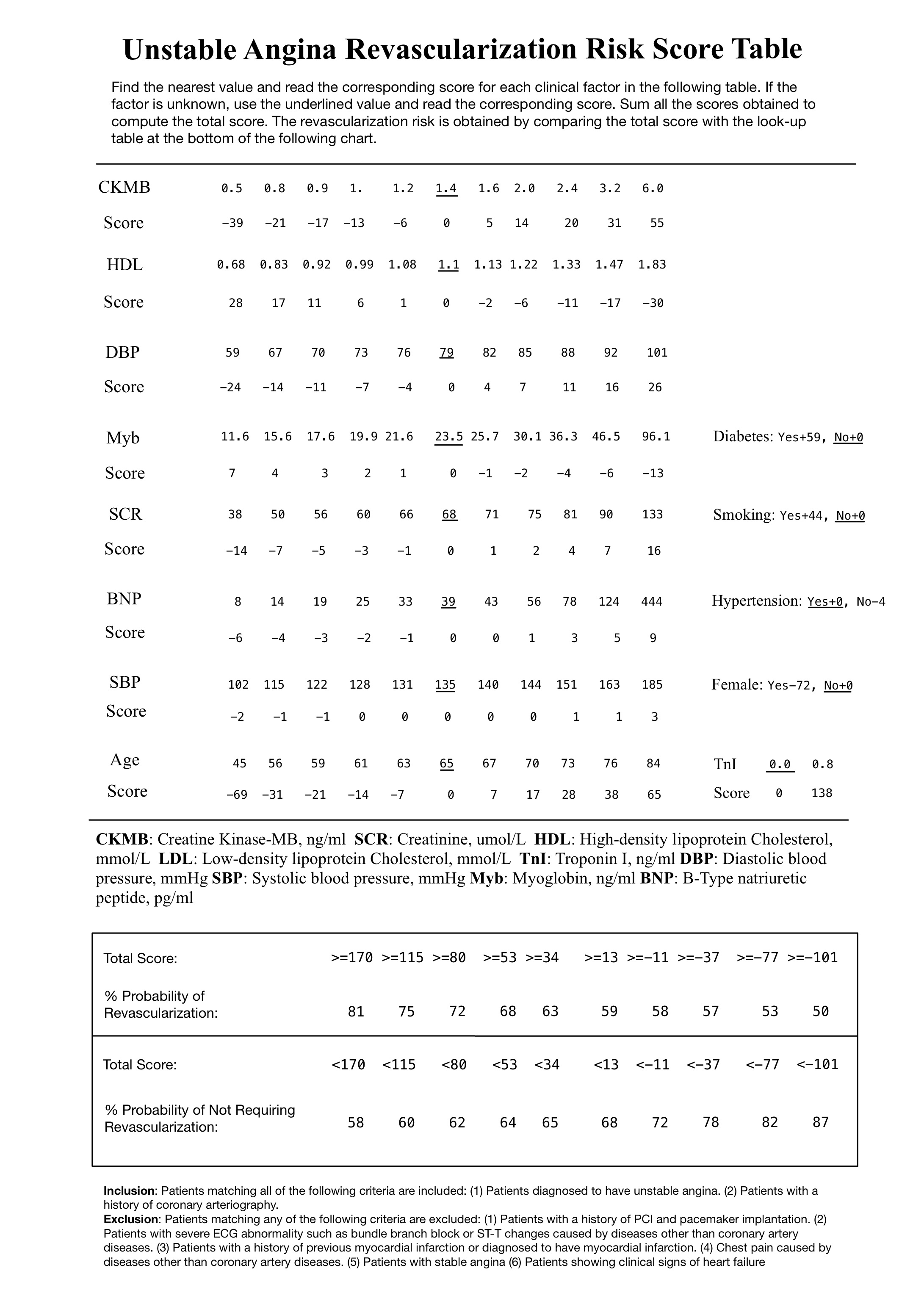}
\end{center}
\caption{Risk scores look-up table Derived from the logistic regression model}\label{fig:3t}
\end{figure}
\paragraph{Declaration of generative AI and AI-assisted technologies in the writing process} During the preparation of this work the author(s) used ChatGPT for grammar checking. After using this tool/service, the author(s) reviewed and edited the content as needed and take(s) full responsibility for the content of the publication.

\bibliographystyle{elsarticle-num} 
\bibliography{refer}








\end{document}